\documentclass[10pt, conference, compsocconf]{IEEEtran}
\IEEEoverridecommandlockouts
\usepackage{cite}
\usepackage{amsmath,amssymb,amsfonts}
\usepackage{textcomp}
\usepackage{amsmath}      
\usepackage{amssymb}      
\usepackage{amsfonts}     
\usepackage{algorithm}
\usepackage{algpseudocode}
\usepackage{graphicx}     
\usepackage{subcaption}   
\usepackage{booktabs}     
\usepackage{multirow}     
\usepackage{xcolor}       
\usepackage{cite}         
\usepackage{xcolor}
\usepackage{soul}
\usepackage{multirow}
\usepackage{graphicx}  

\def\BibTeX{{\rm B\kern-.05em{\sc i\kern-.025em b}\kern-.08em
    T\kern-.1667em\lower.7ex\hbox{E}\kern-.125emX}}
\begin{document}

\title{SPECTRA: An Efficient Spectral-Informed Neural Network for Sensor-Based Activity Recognition\\
\thanks{This work is supported the BMBF in the project CrossAct (01IW25001).}
}

\author{
\IEEEauthorblockN{
Deepika Gurung, Lala Shakti Swarup Ray, Mengxi Liu, Bo Zhou, Paul Lukowicz
}
\IEEEauthorblockA{
\textit{RPTU \& DFKI}\\
Kaiserslautern, Germany\\
deepika.gurung@dfki.de, lala\_shakti\_swarup.ray@dfki.de, mengxi.liu@dfki.de, bo.zhou@dfki.de, paul.lukowicz@dfki.de
}
}

\maketitle

\begin{abstract}
Real-time sensor-based applications in pervasive computing require edge-deployable models to ensure low latency, privacy, and efficient interaction. A prime example is sensor-based human activity recognition (HAR), where models must balance accuracy with stringent resource constraints. Yet many deep learning approaches treat temporal sensor signals as black-box sequences, overlooking spectral--temporal structure while demanding excessive computation. We present SPECTRA, {a deployment-first, co-designed spectral--temporal architecture} that integrates short-time Fourier transform (STFT) feature extraction, depthwise separable convolutions, and channel-wise self-attention to capture spectral--temporal dependencies {under real edge runtime and memory constraints}. A compact bidirectional GRU with attention pooling summarizes within-window dynamics at low cost, {reducing downstream model burden while preserving accuracy}. Across five public HAR datasets, SPECTRA matches or approaches larger CNN/LSTM/Transformer baselines while substantially reducing parameters, latency, and energy. Deployments on a Google Pixel 9 smartphone and an STM32L4 microcontroller further demonstrate {end-to-end deployable} real-time, private, and efficient HAR.
\end{abstract}


\section{Introduction}

The rise of real-time sensor-based applications is reshaping pervasive computing. From health monitoring and rehabilitation to mobility tracking, smart home interaction, and context-aware services, sensors embedded in smartphones and wearables are enabling seamless integration of computing into daily life \cite{Lara2013,bian2022state,Wang2019,Machado2019}. Among these applications, human activity recognition (HAR) using inertial measurement units (IMUs) has become particularly central. IMUs are attractive due to their ubiquity, low cost, and energy efficiency, making them one of the most practical sensing modalities for pervasive systems \cite{Anguita2013}.
Despite their promise, deploying HAR models on pervasive devices remains challenging. Deep learning approaches  spanning convolutional neural networks (CNNs), recurrent networks such as LSTMs and GRUs, and more recent Transformer-based models  have delivered strong accuracy in controlled research settings \cite{Ravi2016,Wang2019}. Yet these models are computationally intensive, often with millions of parameters and high memory demands, making them unsuitable for smartphones, wearables, and microcontrollers with limited resources \cite{Lane2015}. Offloading sensor streams to the cloud is not a sustainable alternative due to privacy, connectivity, and latency concerns \cite{Ravi2016,liu2024wearable}.  
A more fundamental limitation lies in how most HAR models treat IMU signals: as black-box temporal sequences. While recurrent and attention-based models capture temporal dependencies, they generally overlook the {spectral temporal structure} that underpins human motion. Frequency-domain cues such as cadence in walking, harmonic patterns in running, or subtle vibration dynamics in fine-grained gestures are highly discriminative yet rarely exploited in lightweight HAR architectures. Incorporating such information efficiently could enable higher recognition quality under strict resource constraints.
To address these gaps we propose SPECTRA, {a deployment-first, co-designed} spectral-informed neural architecture for on-device HAR that is explicitly designed to balance accuracy and efficiency. {Our contribution is not any single new operator, but the systematic integration of spectral inductive bias with lightweight temporal modeling under real edge constraints (latency, memory, energy, and operator/toolchain support).}
SPECTRA integrates a short-time Fourier transform (STFT) front-end with depthwise separable convolutions and channel-wise self-attention to efficiently capture both spectral patterns and cross-sensor dependencies. Temporal dynamics are modeled using a compact bidirectional GRU with attention pooling, chosen for its balance of expressive power and low complexity compared to heavier LSTM or Transformer alternatives. The result is a compact yet expressive model architecture suitable for pervasive devices {because the spectral front-end reduces downstream model burden}.
We validate SPECTRA extensively through experiments and real deployments. Evaluations on five public HAR datasets show that our approach achieves accuracy comparable to larger CNN, LSTM, and Transformer baselines, while reducing parameters, latency, and energy by large margins. Real-time deployments on a Google Pixel 9 smartphone (ExecuTorch, float32) and an STM32L4S5VIT6 microcontroller (ONNX $\rightarrow$ TFLM, float32) confirm its practicality for embedded, privacy-preserving pervasive sensing {and demonstrate end-to-end feasibility under realistic runtime and operator constraints}.

In summary, this paper makes the following contributions:
\begin{itemize}
    \item {We present SPECTRA as a deployment-first co-design of model and inference pipeline for edge HAR, integrating spectral inductive bias (STFT) with compact temporal modeling to optimize accuracy--latency--energy--memory trade-offs, and} demonstrate end-to-end deployment on both smartphone and microcontroller hardware, supporting efficient inference pipelines tailored to edge platforms.
    \item We evaluate extensively on five public HAR datasets, showing that SPECTRA matches or exceeds larger models in accuracy while drastically lowering computational and energy costs.
\end{itemize}

\section{Related Work}

\subsection{From Classical Approaches to Deep Learning for HAR}
Early IMU-based HAR relied on handcrafted statistical (and sometimes spectral) features computed over fixed-length windows, e.g., mean, variance, and entropy, classified with SVM, RF, or kNN~\cite{Lara2013, Anguita2013}. While effective in controlled settings, such pipelines hinge on expert feature design and often generalize poorly across activities, placements, and devices~\cite{Wang2019}.  
Deep learning shifted HAR toward end-to-end representation learning. CNNs learn discriminative patterns directly from raw signals~\cite{Ronao2016}, while RNNs (LSTMs, GRUs) capture temporal dependencies~\cite{Murad2017}. Hybrid CNN RNN(LSTM) models further improved accuracy~\cite{ordonez2016deep, Yao2018}, and attention/Transformer architectures explored long-range relations~\cite{Vaswani2017, ray2024har}. However, these models are computationally and memory intensive, limiting on-device use in pervasive settings~\cite{Lane2015 ,ray2025improving ,fortes2024enhancing }.

\subsection{Frequency and Spectral Approaches in Deep Learning}
Frequency-domain representations expose periodicities, rhythmic structures, and localized spectral patterns less apparent in time-domain signals. Classical signal-processing and ML applications like speech/audio recognition and biomedical analysis routinely leverage Fourier and wavelet transforms for discriminative structure.  
Within deep learning spectral features are widely integrated into neural models for speech recognition, audio event detection, and EEG-based interfaces, enabling networks to learn temporal spectral dependencies rather than purely temporal correlations.  
For IMU-based HAR, transforming accelerometer/gyroscope signals into spectrograms or wavelet coefficients helps capture activity-specific rhythmic signatures ~\cite{zhou2016smart,Hassan2018, ignatov2018real, Gao2020}. STFT features, in particular, provide explicit spectral temporal structure that can bolster robustness and reduce reliance on very deep CNN stacks~\cite{Hassan2018}. Nevertheless, many spectral pipelines still employ large backbones, leaving deployability on resource-constrained devices unresolved.

\subsection{Efficient and Edge-oriented HAR}
A complementary line of work targets resource efficiency. Pruning, quantization, and distillation shrink models while attempting to preserve accuracy~\cite{Han2015, Jacob2018, Machado2019}. Depthwise separable convolutions reduce redundancy in convolutional layers and underpin many mobile architectures~\cite{Howard2017}. Despite notable savings in memory and energy, efficient HAR models often trade away accuracy or remain validated only in simulation rather than on real devices~\cite{Ravi2016}.  
Bridging these strands, we pursue a spectral-informed yet lightweight design tailored for edge deployment. Unlike prior spectral works that rely on heavy backbones, and unlike purely efficiency-driven models that degrade accuracy, our approach integrates an STFT front-end with depthwise separable convolutions, channel-wise self-attention, and a compact Bi-GRU to jointly preserve spectral richness and achieve on-device efficiency across heterogeneous platforms~\cite{Agarwal2020, liu2024wearable}.

\section{Model Architecture}

\begin{figure}[t]
  \centering
 \includegraphics[width=1\linewidth]{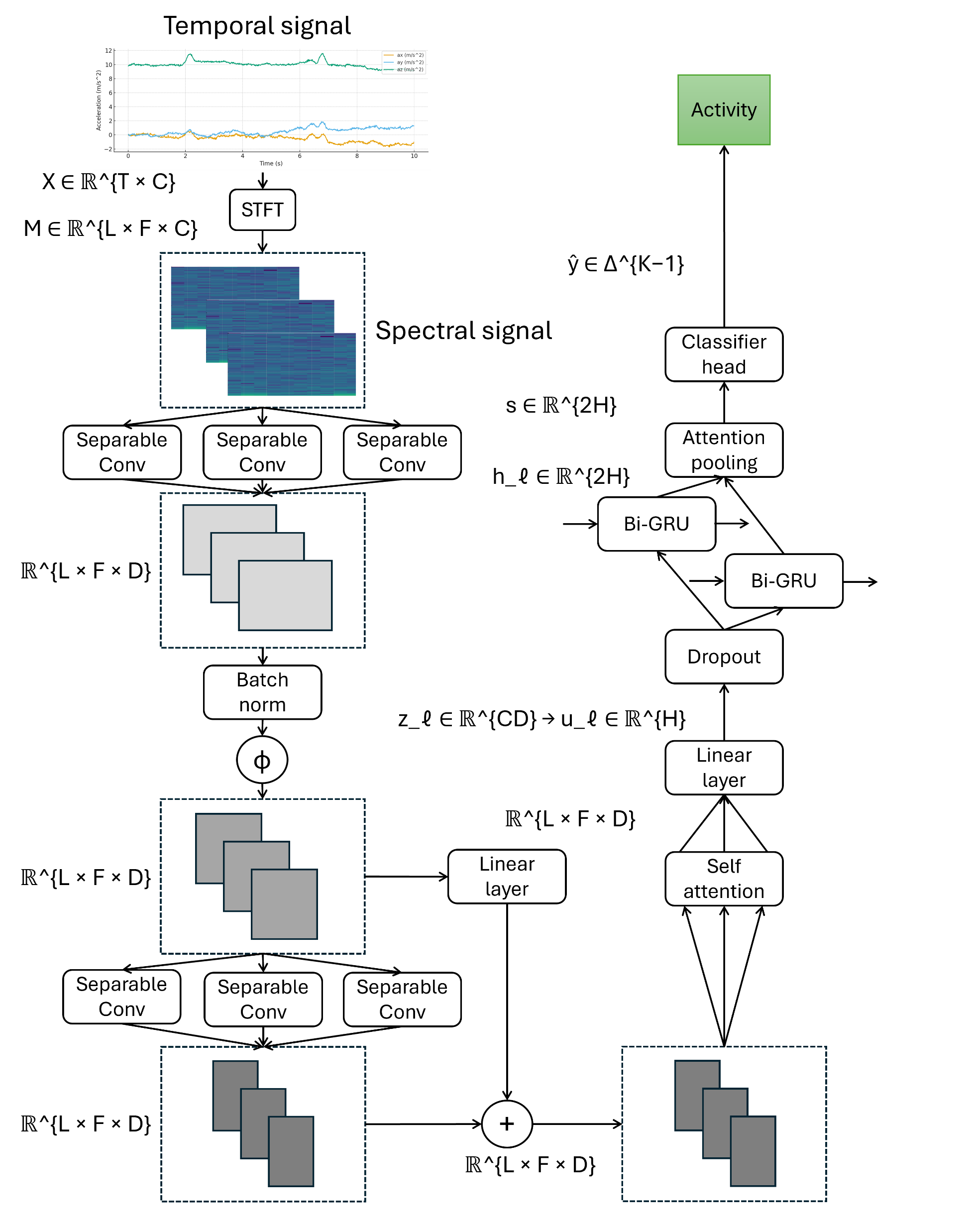}
\caption{SPECTRA architecture: signal-informed spectral--temporal modeling achieves competitive HAR accuracy with a compact, shallow pipeline.
\protect{Legend: input window $X\in \mathrm{R}^{T\times C}$ ($T$ samples/window, $C$ IMU channels/axes, dataset-dependent); STFT magnitudes $M\in \mathrm{R}^{L\times F\times C}$ ($L$ STFT frames, $F = n_{\mathrm{fft}}/2 + 1$ freq bins); separable conv maps to $L\times F\times D$ features ($D$ conv feature dim); channel self-attention operates over $C$ channels per frame; Bi-GRU with hidden size $H$ and attention pooling aggregates over $L$ to a $2H$ vector; classifier outputs $\hat{\mathbf{y}}$ over $K$ activities.}}

  \label{fig:spectra-vs-sota}
\end{figure}

The proposed SPECTRA model balances recognition accuracy with computational efficiency, enabling deployment on smartphones and microcontrollers. 
Figure~\ref{fig:spectra-vs-sota} shows the end-to-end pipeline:
Let the windowed IMU input be \(X \in \mathbb{R}^{T \times C}\), where \(T\) is the number of samples per window and \(C\) the number of sensor channels. After processing, the model outputs class probabilities \(\hat{\mathbf{y}} \in \Delta^{K-1}\) over \(K\) activities.
{ SPECTRA follows a deployment-first co-design principle: \textit{front-load} domain-aligned time--frequency structure to reduce downstream model burden, then allocate lightweight capacity to (i) local motif extraction, (ii) cross-channel coupling, and (iii) within-window temporal dynamics. Concretely, \textbf{STFT} exposes nonstationary spectral cues with a fixed, controllable cost; \textbf{separable convolutions} efficiently capture local spectral--temporal motifs with minimal parameters/MACs; \textbf{channel self-attention} selectively fuses IMU axes/sensors where inter-axis coupling is discriminative, at low overhead because $C$ is small; and \textbf{Bi-GRU with attention pooling} provides just-enough temporal modeling and frame selection without the quadratic cost and memory of Transformers.}

\subsection{STFT Feature Extraction}
We segment each IMU stream into overlapping windows of length $T$ with hop $h$, and compute a per channel short-time Fourier transform (STFT) using a Hann window $w[\cdot]$, FFT size $n_{\mathrm{fft}}$, and hop $h$. For channel $c$ and frame $\ell=0,\dots,L-1$,
\begin{equation}
\begin{aligned}
\mathrm{STFT}_c[\ell,f]
&= \sum_{t=0}^{T-1} X[t,c]\; w[t-\ell h]\; e^{-j2\pi ft/n_{\mathrm{fft}}}, \\
&\quad f=0,\dots,F-1.
\end{aligned}
\end{equation}
\begin{equation}
F=\frac{n_{\mathrm{fft}}}{2}+1.
\end{equation}
We retain magnitudes and stack across channels,
\begin{equation}
M[\ell,f,c] = \big|\mathrm{STFT}_c[\ell,f]\big|,\qquad
M \in \mathbb{R}^{L\times F\times C}.
\end{equation}

Unlike a global FFT, the STFT preserves {time--frequency} localization for nonstationary motion. Using magnitudes only keeps the front end compact and device-friendly while capturing discriminative spectral patterns. The computational cost is $O(L\,F\log F)$, and the pair $(n_{\mathrm{fft}},h)$ offers a tunable accuracy--latency trade-off suitable for mobile/MCU deployment.
{\noindent\textit{Efficiency intuition:} STFT provides an explicit, hardware-friendly inductive bias that would otherwise require a much deeper (and costlier) raw-signal network to approximate; our ablations show that removing STFT both reduces accuracy and forces a large increase in downstream compute to recover performance.}

\subsection{Convolutional Feature Block}
We apply a depthwise separable 2D convolution across the \((\ell, f)\) axes with per-channel filters followed by a pointwise projection. 
Let the block input be \(U \in \mathbb{R}^{L \times F \times C}\). 
A depthwise conv with temporal kernel \(k\) (along \(L\)) acts independently on each channel feature map:
\begin{equation}
\tilde{U}[:,:,c] \;=\; U[:,:,c] \;*\; \Theta_c \;\;\;\; \forall c \in \{1,\dots,C\},
\end{equation}
followed by a pointwise \(1{\times}1\) conv that mixes channels into \(D\) feature maps:
\(
V = \mathrm{Conv}_{1\times 1}(\tilde{U}) \in \mathbb{R}^{L \times F \times D}.
\)
We use BatchNorm and ReLU and add a residual shortcut if dimensions match. 

We want local pattern extraction at edge cost. A standard \(k{\times}1\) conv (mapping \(C{\to}D\)) has \(kCD\) params; separable has \(kC + CD\). The reduction factor
\begin{equation}
\frac{kCD}{kC + CD} \gg 1 \quad (\text{for } D \gg k)
\end{equation}
yields order-of-magnitude savings in params/MACs and better cache locality critical for real-time on MCUs.
{\noindent\textit{Functional role:} this block learns compact local ``motifs'' in the time--frequency plane while keeping the receptive field and compute predictable for edge deployment.}

\subsection{Channel Self-Attention}
Concatenation treats channels independently. We explicitly model cross-channel dependencies at each frame \(\ell\). Let
\(
X_\ell \in \mathbb{R}^{C \times D}
\)
be the per-frame feature matrix after the conv block (rows=channels, cols=features). We learn channel-space queries, keys, and values via shared linear maps
\(
W_q,W_k,W_v \in \mathbb{R}^{D \times D}
\):
\begin{equation}
Q_\ell = X_\ell W_q,\quad K_\ell = X_\ell W_k,\quad V_\ell = X_\ell W_v \quad (\in \mathbb{R}^{C \times D})
\end{equation}
Channel affinities are
\begin{equation}
A_\ell \;=\; \mathrm{softmax}\!\left(\frac{Q_\ell K_\ell^\top}{\sqrt{D}}\right) \in \mathbb{R}^{C \times C}
\end{equation}
and the attended output is
\begin{equation}
\widetilde{X}_\ell \;=\; X_\ell \;+\; \gamma \, A_\ell V_\ell
\end{equation}
with a learnable scalar \(\gamma\). This highlights informative channels while suppressing redundancy. The per-frame cost is \(O(C^2 D)\), but \(C\) is small, so the module remains lightweight.
{\noindent\textit{Efficiency intuition:} channel attention targets the dominant source of redundancy in IMU sensing (correlated axes/sensors) without incurring the $O(L^2)$ cost of temporal self-attention; since $C$ is typically 3--6, the additional compute is negligible on phones and MCUs.}

IMU axes are correlated but not redundant. We explicitly model cross-axis coupling with
\(
Q=XW_q,\, K=XW_k,\, V=XW_v
\),
\(
A=\mathrm{softmax}(QK^\top/\sqrt{D})
\),
\(
\widetilde{X}=X + \gamma AV
\).
This upweights informative axes and downweights noisy ones. Because \(C\) is small (3 6), the \(O(C^2D)\) term is negligible relative to linear projections.

\subsection{Temporal Modeling with Bi-GRU and Attention Pooling}
For sequential modeling across frames, we first vectorize each \(\widetilde{X}_\ell\) as \(\mathbf{z}_\ell \in \mathbb{R}^{C D}\) and project to \(H\) via a learned linear map \(P \in \mathbb{R}^{CD \times H}\):
\begin{equation}
\mathbf{u}_\ell \;=\; \mathbf{z}_\ell P \in \mathbb{R}^{H}
\end{equation}
We then feed \(\{\mathbf{u}_\ell\}_{\ell=1}^L\) to a bidirectional GRU with hidden size \(H\), obtaining \(\mathbf{h}_\ell \in \mathbb{R}^{2H}\). To compress the temporal sequence, we use attention pooling:
\begin{equation}
\begin{aligned}
e_\ell      &= \mathbf{w}^\top \mathbf{h}_\ell,\\
\alpha_\ell &= \frac{\exp(e_\ell)}{\sum_{t=1}^L \exp(e_t)}, \qquad \sum_{\ell=1}^L \alpha_\ell = 1,\\
\mathbf{s}  &= \sum_{\ell=1}^L \alpha_\ell \mathbf{h}_\ell, \qquad \mathbf{s}\in\mathbb{R}^{2H}.
\end{aligned}
\end{equation}

where \(\mathbf{w} \in \mathbb{R}^{2H}\) is learned.
GRU has fewer gates/states than LSTM \(\Rightarrow\) fewer params, lower SRAM bandwidth, faster inference with similar accuracy on short/medium HAR windows. Bidirectionality improves within-window context at minimal cost. Transformers incur \(O(L^2)\) memory/compute and prefer large hidden sizes  unfriendly to MCUs. TCNs need deeper/dilated stacks to match receptive fields, growing params and being sensitive to sampling/window changes. We use attention pooling
\(
e_t=\mathbf{w}^\top\mathbf{h}_t,\,
\alpha_t=\mathrm{softmax}(e_t),\,
\mathbf{s}=\sum_t \alpha_t \mathbf{h}_t
\)
to emphasize informative frames with negligible overhead.
{\noindent\textit{Functional role:} the Bi-GRU provides low-cost temporal smoothing and ordering information across STFT frames, while attention pooling acts as a lightweight frame selector that emphasizes salient motion events without adding heavy sequence-level compute.}

\subsection{Classifier and Training Objective}
We apply dropout to \(\mathbf{s}\) and a linear classifier:
\begin{equation}
\hat{\mathbf{y}} \;=\; \mathrm{softmax}(W_c \mathbf{s} + \mathbf{b}_c) \in \mathbb{R}^{K}
\end{equation}
The model is trained with
\(
\mathcal{L} \;=\; - \sum_{k=1}^{K} y_k \log \hat{y}_k,
\)
where \(\mathbf{y}\) is the one-hot label. All blocks are differentiable; gradients propagate from classifier to STFT front-end through the convolutional and attention layers.
{Importantly, SPECTRA is co-designed with deployment constraints in mind, so the reported gains reflect end-to-end deployable pipelines rather than simulation-only efficiency.}

\section{Deployment Optimization}

A core design goal of SPECTRA is to ensure practical feasibility across heterogeneous pervasive platforms. To this end, we established two deployment pipelines: one for modern smartphones and one for resource-constrained microcontrollers. In this section we describe the toolchains, conversion steps, and execution settings used to prepare the model for deployment. Actual performance results are presented later in Section~\ref{sec:eval}.

\subsection{Smartphone Deployment: Google Pixel 9}
We deployed all models on a Google Pixel 9 (Android 14; Google Tensor G4 SoC; 16\,GB LPDDR5X) using ExecuTorch. Models were trained in PyTorch, exported to TorchScript, and integrated into an Android app (Android Studio), executing via the NNAPI backend with batch size $B{=}1$ and fixed HAR windowing ($T{=}2$\,s, $C{=}6$). Because ExecuTorch does not support a native STFT operator, SPECTRA’s STFT front-end was implemented as a bank of 1D convolutional filters initialized as Hann-windowed complex exponentials, which exactly reproduces STFT computation while remaining runtime-compatible. This yields an end-to-end on-device evaluation under realistic latency, memory, and energy constraints without modifying the SPECTRA architecture.

\subsection{Microcontroller Deployment: STM32L4S5VIT6}
For embedded deployment, we targeted the STM32L4S5VIT6 (ARM Cortex-M4, 80\,MHz; 2\,MB Flash; 640\,KiB RAM on NUCLEO board) using an ONNX $\rightarrow$ TFLite $\rightarrow$ TFLM toolchain with CMSIS-NN acceleration. Models were exported from PyTorch to ONNX, converted to TFLM, and executed with $B{=}1$ and the same $T{=}2$\,s windowing for comparability. Latency/throughput/energy in Table~\ref{tab:mcu_results} were measured on-device, with power profiled externally on the NUCLEO-L4R5ZI; Params/MACs were computed from the exported ONNX graph, and training results are reported over three seeds. Due to TFLM operator limitations, recurrent layers were replaced with lightweight 1D convolutional blocks (SPECTRA: Bi-GRU; DeepConvLSTM/TinyHAR: LSTM) to enable execution, while TinierHAR could not be deployed because it relies on unsupported operators with no faithful workaround.

\section{Experimental Evaluation}
\label{sec:eval}
We evaluate SPECTRA on five public IMU-based HAR datasets and across three deployment scenarios: (i) PC workstation for reference training and benchmarking, (ii) smartphone deployment on a Google Pixel 9, and (iii) microcontroller deployment on STM32L4S5VIT6. The evaluation examines classification performance, computational efficiency, and on-device feasibility.

{Unless otherwise stated, all baseline methods (DeepConvLSTM, TinyHAR, TinierHAR) are evaluated in FP32. For the proposed method, we evaluate three inference variants: SPECTRA (FP32), SPECTRA (FP16), and SPECTRA (INT8). These variants share the same architecture; only the numeric precision used during inference changes.}

We use five widely adopted HAR datasets \textbf{WISDM} \cite{weiss2019wisdm}, \textbf{USC-HAD} \cite{zhang2012usc}, \textbf{UCI HAR} \cite{ignatov2018real}, \textbf{DSADS} and  \textbf{PAMAP2} \cite{reiss2012introducing}. These datasets span a range of activity types and sensing contexts, ensuring robust evaluation across domains.

All datasets were resampled to a uniform frequency of 50\,Hz when necessary. Signals were segmented into fixed-length windows of $T=2$\,s with 50\% overlap. Each channel was normalized to zero mean and unit variance. Input tensors have the form $(B, 1, T, C)$, where $C=6$ for accelerometer and gyroscope channels unless otherwise specified.

We compare SPECTRA with three representative HAR baselines like \textbf{DeepConvLSTM}~\cite{ordonez2016deep}   a widely used CNN+LSTM hybrid model, considered a strong benchmark in HAR. \textbf{TinyHAR}~\cite{zhou2022tinyhar}   a lightweight CNN-based HAR model specifically optimized for microcontrollers. \textbf{TinerHAR}~\cite{bian2025tinierhar}   a further optimized variant of TinyHAR with reduced memory footprint and computation cost.
This baseline set allows us to compare SPECTRA against both accuracy-focused (DeepConvLSTM) and efficiency-focused (TinyHAR/TinerHAR) state-of-the-art methods.

We report Macro F1-score ($\pm$ std) and accuracy, parameter count (K) and multiply accumulate operations (MACs), latency per window (s), throughput (samples/s), memory footprint (MB), and energy cost ( J, J/sample).

{ On the workstation, FP16 uses half-precision execution for supported operators to reduce memory bandwidth and improve throughput. INT8 uses the (Post training quantization) PTQ-exported quantized model; operators with available INT8 kernels execute in INT8, and any unsupported operators fall back to higher precision through dequantization, preserving correctness while still reducing cost where INT8 kernels are available.}
{On Pixel~9, that uses ExecuTorch with the Android NNAPI backend, FP32 runs with float32 operators. FP16 uses NNAPI float16 execution for supported layers (reduced-precision arithmetic and reduced activation/weight bandwidth). INT8 uses an NNAPI quantized PTQ model where weights/activations are quantized to INT8 using calibration-derived scales/zero-points; NNAPI executes supported quantized operators with device-accelerated kernels. When NNAPI does not support a specific operator or configuration, ExecuTorch falls back to CPU kernels (e.g., XNNPACK) for that portion of the graph, potentially in higher precision, while keeping the overall model output identical.}
{ On STM32, that uses (ONNX$\rightarrow$TFLite) to run it in TFLM with CMSIS-NN acceleration, FP32 uses float32 TFLM kernels. INT8 uses a fully-quantized INT8 TFLite graph produced by PTQ calibration so that supported layers map to CMSIS-NN INT8 kernels; float16 is not supported in this toolchain/hardware, so we do not report FP16 on STM32.}

\subsection{Results on PC}
Table~\ref{tab:pc_results} reports results across five benchmark datasets on a workstation GPU (NVIDIA RTX A6000 Ada Lovelace, 48\,GB). {Parameter counts differ slightly across datasets because the input dimensionality (IMU channels/axes) and resulting spectral-bin dimensionality affect the first projection layers; the core SPECTRA blocks (separable conv, channel attention, and Bi-GRU) are unchanged.} SPECTRA provides a strong accuracy--efficiency trade-off: in FP32 it achieves the best F1-score on 2/5 datasets (USC-HAD, DSADS), second-best on 2 (WISDM, PAMAP2), and stays within 0.7\% of DeepConvLSTM on UCI HAR, while being dramatically smaller and cheaper (97--99\% fewer parameters and 99\%+ fewer MACs vs.\ DeepConvLSTM; 65--75\% fewer parameters and 85--95\% fewer MACs vs.\ TinyHAR, and still substantially fewer MACs than TinierHAR). {\noindent\textbf{End-to-end cost accounting.} All runtime/energy numbers in Table~}\ref{tab:pc_results}{ are end-to-end and include STFT feature extraction; for SPECTRA variants we report latency explicitly as }$t_{\mathrm{stft}}{+}t_{\mathrm{nn}}${ (ms), whereas other models report a single total latency.} Concretely, FP32 SPECTRA runs in 1.48--3.21\,ms end-to-end and delivers 353.7--671.1\,samples/s, while also achieving the lowest energy per sample and the smallest peak memory among compared models (26.9--120.7\,MB). {Reduced precision further strengthens deployability without changing the architecture: FP16/INT8 keep the same parameters and MACs but reduce end-to-end latency/energy and memory (peak memory drops to roughly 17.5--18.7\,MB), with FP16 reaching 1502.5--1875.6\,samples/s and INT8 reaching 1040.3--1241.2\,samples/s; INT8 yields the lowest energy per sample, while FP16 is often the lowest in total Joules on this GPU setup.} Overall, SPECTRA attains top-tier accuracy while delivering the best combined latency, throughput, memory, and energy efficiency on PC, {and FP16/INT8 variants further improve end-to-end deployability under identical model structure.}

\begin{table*}[ht]
\centering
\caption{Results on PC workstation across five datasets. 
Metrics reported: F1-score ($\uparrow$), parameters (K, $\downarrow$), multiply accumulate operations (MACs, $\downarrow$), 
inference latency (ms, $\downarrow$), throughput (samples/s, $\uparrow$), 
peak memory (MEM, $\downarrow$), total energy (J, $\downarrow$), and energy per sample (J/sample, $\downarrow$). 
{For SPECTRA variants, latency is reported as STFT feature transformation time $+$ neural backbone time (ms); other models report directly total latency.}
Best results are in \textbf{bold}, second best are \underline{underlined}.}
\label{tab:pc_results}
\begin{tabular}{lccccccccc}
\toprule
\textbf{Dataset} & \textbf{Model} & \textbf{F1 $\pm$ std (\%)} ($\uparrow$) & \textbf{K} ($\downarrow$) & \textbf{MACs} ($\downarrow$) & {\textbf{ms ($\downarrow$)}} & \textbf{samples/s} ($\uparrow$) & \textbf{Mem} ($\downarrow$) & \textbf{J} ($\downarrow$) & \textbf{J/sample} ($\downarrow$) \\
\midrule

\multirow{6}{*}{WISDM} 
 & DeepConvLSTM & 83.2 $\pm$ 0.3 & 227654 & 9940678 & 10.34 & 96.7 & 76.7 & 150.3 & 0.00587 \\
 & TinyHAR & \textbf{84.1} $\pm$ \textbf{0.3} & 25028 & 1144056 & 4.70 & 212.8 & \underline{33.9} & 113.4 & 0.00443 \\
 & TinierHAR & 82.1 $\pm$ 0.6 & \textbf{4981} & \underline{199223} & \underline{2.48} & \underline{403.2} & 39.8 & \underline{46.7} & \underline{0.00183} \\
 & {SPECTRA (FP32)} & \underline{83.6} $\pm$ \underline{0.5} & \underline{8892} & \textbf{28530} & {$0.52{+}0.96$} & \textbf{671.1} & \textbf{26.9} & \textbf{12.1} & \textbf{0.00047} \\
 & {SPECTRA (FP16)} & {82.7 $\pm$ 0.4} & 8892 & 28530 & {$0.18{+}0.35$} & {1850.4} & {18.7} & {0.726} & {0.00007} \\
 & {SPECTRA (INT8)} & {82.0 $\pm$ 0.6} & 8892 & 28530 & {$0.28{+}0.52$} & {1241.2} & {18.7} & {1.211} & {0.00012} \\
\midrule

\multirow{6}{*}{USC-HAD} 
 & DeepConvLSTM & \underline{71.1} $\pm$ \underline{0.2} & 326732 & 18592396 & 18.69 & 53.5 & 124.1 & 272.4 & 0.0106 \\
 & TinyHAR & 68.4 $\pm$ 1.3 & 27674 & 2011035 & 6.98 & 143.3 & 55.5 & 145.5 & 0.00568 \\
 & TinierHAR & 70.3 $\pm$ 0.5 & \textbf{7483} & \underline{348821} & \textbf{1.83} & \textbf{546.4} & \underline{44.0} & \underline{48.2} & \underline{0.00188} \\
 & {SPECTRA (FP32)} & \textbf{71.2} $\pm$ \textbf{0.7} & \underline{10050} & \textbf{51462} & {$0.69{+}1.28$} & \underline{507.6} & \textbf{27.5} & \textbf{11.4} & \textbf{0.00044} \\
 & {SPECTRA (FP16)} & {71.0 $\pm$ 0.6} & 10050 & 51462 & {$0.18{+}0.34$} & {1875.6} & {17.6} & {0.765} & {0.00007} \\
 & {SPECTRA (INT8)} & {69.4 $\pm$ 0.1} & 10050 & 51462 & {$0.31{+}0.56$} & {1144.7} & {17.6} & {1.022} & {0.00010} \\
\midrule

\multirow{6}{*}{UCI HAR} 
 & DeepConvLSTM & \textbf{95.5} $\pm$ \textbf{0.7} & 424262 & 36422534 & 35.55 & 28.1 & 226.9 & 554.3 & 0.0217 \\
 & TinyHAR & \underline{95.3} $\pm$ \underline{0.2} & 29828 & 3859938 & 11.71 & 85.4 & 99.6 & 199.9 & 0.00781 \\
 & TinierHAR & 95.1 $\pm$ 0.6 & \textbf{9589} & \underline{637414} & \textbf{2.68} & \textbf{373.1} & \underline{55.4} & \underline{60.3} & \underline{0.00235} \\
 & {SPECTRA (FP32)} & 94.8 $\pm$ 0.2 & \underline{10812} & \textbf{113112} & {$0.99{+}1.84$} & \underline{353.7} & \textbf{38.2} & \textbf{29.5} & \textbf{0.00115} \\
 & {SPECTRA (FP16)} & {93.3 $\pm$ 0.5} & 10812 & 113112 & {$0.21{+}0.38$} & {1677.5} & {17.6} & {0.955} & {0.00009} \\
 & {SPECTRA (INT8)} & {85.5 $\pm$ 0.3} & 10812 & 113112 & {$0.33{+}0.62$} & {1040.3} & {17.6} & {1.238} & {0.00012} \\
\midrule

\multirow{6}{*}{DSADS} 
 & DeepConvLSTM & 84.8 $\pm$ 0.4 & 1605587 & 173126291 & 166.34 & 6.01 & 1044.2 & 2732.5 & 0.107 \\
 & TinyHAR & \underline{86.4} $\pm$ \underline{0.8} & 59161 & 19216775 & 51.51 & 19.4 & 445.2 & 819.7 & 0.0320 \\
 & TinierHAR & 86.2 $\pm$ 0.3 & \underline{ } & \underline{2843792} & \underline{4.37} & \underline{229.1} & \underline{150.6} & \underline{148.1} & \underline{0.00579} \\
 & {SPECTRA (FP32)} & \textbf{87.6} $\pm$ \textbf{0.2} & \textbf{22761} & \textbf{481551} & {$1.16{+}2.05$} & \textbf{316.5} & \textbf{120.7} & \textbf{46.5} & \textbf{0.00182} \\
 & {SPECTRA (FP16)} & {86.9 $\pm$ 0.5} & 22761 & 481551 & {$0.27{+}0.38$} & {1687.7} & {17.6} & {0.944} & {0.00009} \\
 & {SPECTRA (INT8)} & {85.7 $\pm$ 0.3} & 22761 & 481551 & {$0.28{+}0.51$} & {1258.7} & {17.5} & {1.018} & {0.00010} \\
\midrule

\multirow{6}{*}{PAMAP2} 
 & DeepConvLSTM & 72.6 $\pm$ 0.6 & 719948 & 96631308 & 93.39 & 10.7 & 571.6 & 1541.9 & 0.0602 \\
 & TinyHAR & 73.7 $\pm$ 1.1 & 37274 & 10174272 & 29.03 & 34.5 & 246.3 & 437.6 & 0.0171 \\
 & TinierHAR & \textbf{75.6} $\pm$ \textbf{2.2} & \underline{16699} & \underline{1589718} & \underline{3.13} & \underline{319.5} & \underline{85.4} & \underline{91.7} & \underline{0.00358} \\
 & {SPECTRA (FP32)} & \underline{75.0} $\pm$ \underline{0.7} & \textbf{13890} & \textbf{235539} & {$0.72{+}1.33$} & \textbf{485.4} & \textbf{72.4} & \textbf{37.7} & \textbf{0.00147} \\
 & {SPECTRA (FP16)} & {73.6 $\pm$ 0.2} & 13890 & 235539 & {$0.23{+}0.43$} & {1502.5} & {17.7} & {0.867} & {0.00008} \\
 & {SPECTRA (INT8)} & {72.0 $\pm$ 0.4} & 13890 & 235539 & {$0.32{+}0.59$} & {1086.1} & {17.7} & {0.917} & {0.00009} \\
\bottomrule
\end{tabular}
\end{table*}

\subsection{Results on Smartphone (Pixel 9)}
We deployed SPECTRA on a Google Pixel~9 using ExecuTorch in FP32, FP16, and INT8. 
Table~\ref{tab:smartphone_results} reports on-device performance across five datasets, and {all numbers are measured end-to-end on-device and include STFT computation}. 
Across precisions, SPECTRA achieves real-time operation: FP32 latency is 1.7--3.4\,ms (333.3--588.2\,samples/s), FP16 latency is 1.4--2.9\,ms (346.0--692.0\,samples/s), and INT8 latency is 1.2--2.4\,ms (420.2--840.3\,samples/s), while maintaining the smallest footprint and lowest energy among compared models. 
\textbf{Latency \& Throughput.} SPECTRA attains the best latency and throughput on all datasets for all three precisions: in FP32 it is 3.1$\times$--15.3$\times$ faster than TinyHAR and 6.7$\times$--43.8$\times$ faster than DeepConvLSTM, and it remains 1.03$\times$--1.55$\times$ lower latency than TinierHAR; {lower precision further improves speed} (FP16 reduces latency by roughly 15--20\% and INT8 by roughly 25--30\% vs.\ FP32, with matching throughput gains). 
\textbf{Energy.} SPECTRA delivers the lowest energy per sample on all datasets: FP32 uses 4.1--9.8\,mJ/sample (72--92\% lower than TinyHAR, 11--36\% lower than TinierHAR, and 86--97\% lower than DeepConvLSTM), and {reduced precision lowers energy further} (FP16: 3.5--8.3\,mJ/sample; INT8: 2.9--6.9\,mJ/sample), yielding the lowest total energy as well. 
\textbf{Memory footprint.} Peak memory is minimal for SPECTRA in FP32 (27.0--120.8\,KB), about $\sim$2$\times$ smaller than TinierHAR and $\sim$4.5$\times$ smaller than TinyHAR on average, and $\sim$28$\times$--42$\times$ smaller than DeepConvLSTM; {lower precision reduces memory further} (FP16: 24.3--108.7\,KB; INT8: 21.6--96.6\,KB). 
Overall, on Pixel~9 SPECTRA is best across latency, throughput, energy, and memory on all datasets in Table~\ref{tab:smartphone_results}.

\begin{table*}[ht]
\centering
\caption{Deployment results on Google Pixel 9 (ExecuTorch, float32, {float16 and INT8}). 
Metrics reported: parameters (K, $\downarrow$), multiply accumulate operations (MACs, $\downarrow$), 
inference latency (ms, $\downarrow$), throughput (samples/s, $\uparrow$), 
peak memory (KB, $\downarrow$), total energy (mJ, $\downarrow$), and energy per sample (mJ/sample, $\downarrow$). 
Best results are in \textbf{bold}, second best are \underline{underlined}.}
\label{tab:smartphone_results}
\begin{tabular}{lcccccc}
\toprule
\textbf{Dataset} & \textbf{Model} & \textbf{Latency (ms $\downarrow$)} & \textbf{samples/s ($\uparrow$)} & \textbf{Mem (KB $\downarrow$)} & \textbf{Energy (mJ $\downarrow$)} & \textbf{mJ/sample ($\downarrow$)} \\
\midrule
\multirow{6}{*}{WISDM}
 & TinyHAR      & 5.2   & 192.3  & 121.5  & 13{,}838.0 & 14.8 \\
 & TinierHAR    & \underline{2.6} & \underline{384.6} & \underline{54.0} & \underline{5{,}797.0}  & \underline{6.2} \\
 & DConvLSTM    & 11.4  & 87.7   & 762.0  & 27{,}021.5 & 28.9 \\
 & SPECTRA {(FP32)} & \textbf{1.7} & \textbf{588.2} & \textbf{27.0} & \textbf{3{,}833.5} & \textbf{4.1} \\
 & {SPECTRA (FP16)} & {1.4} & {692.0} & {24.3} & {3{,}258.5} & {3.5} \\
 & {SPECTRA (INT8)} & {1.2} & {840.3} & {21.6} & {2{,}683.4} & {2.9} \\
\midrule
\multirow{6}{*}{USC-HAD}
 & TinyHAR      & 7.4   & 135.1  & 123.8  & 11{,}564.0 & 19.6 \\
 & TinierHAR    & \underline{2.2} & \underline{454.5} & \underline{55.0} & \underline{3{,}186.0}  & \underline{5.4} \\
 & DConvLSTM    & 19.3  & 51.8   & 1,165.4 & 24{,}603.0 & 41.7 \\
 & SPECTRA {(FP32)} & \textbf{2.0} & \textbf{500.0} & \textbf{27.5} & \textbf{2{,}832.0} & \textbf{4.8} \\
 & {SPECTRA (FP16)} & {1.7} & {588.2} & {24.8} & {2{,}407.2} & {4.1} \\
 & {SPECTRA (INT8)} & {1.4} & {714.3} & {22.0} & {1{,}982.4} & {3.4} \\
\midrule
\multirow{6}{*}{UCI HAR}
 & TinyHAR      & 12.9  & 77.5   & 173.3  & 9{,}392.2 & 31.1 \\
 & TinierHAR    & \underline{3.1} & \underline{322.6} & \underline{77.0} & \underline{2{,}295.2}  & \underline{7.6} \\
 & DConvLSTM    & 37.5  & 26.7   & 1231.0 & 23{,}978.8 & 79.4 \\
 & SPECTRA {(FP32)} & \textbf{3.0} & \textbf{333.3} & \textbf{38.5} & \textbf{1{,}963.0} & \textbf{6.5} \\
 & {SPECTRA (FP16)} & {2.5} & {392.2} & {34.6} & {1{,}668.5} & {5.5} \\
 & {SPECTRA (INT8)} & {2.1} & {476.2} & {30.8} & {1{,}374.1} & {4.5} \\
\midrule
\multirow{6}{*}{DSADS}
 & TinyHAR      & 52.0  & 19.2   & 543.6  & 12{,}626.0 & 118.0 \\
 & TinierHAR    & \underline{4.6} & \underline{217.4} & \underline{241.6} & \underline{1{,}626.4}  & \underline{15.2} \\
 & DConvLSTM    & 168.0 & 6.0    & 4724.8 & 34{,}775.0 & 325.0 \\
 & SPECTRA {(FP32)} & \textbf{3.4} & \textbf{294.1} & \textbf{120.8} & \textbf{1{,}048.6} & \textbf{9.8} \\
 & {SPECTRA (FP16)} & {2.9} & {346.0} & {108.7} & {891.3} & {8.3} \\
 & {SPECTRA (INT8)} & {2.4} & {420.2} & {96.6} & {734.0} & {6.9} \\
\midrule
\multirow{6}{*}{PAMAP2}
 & TinyHAR      & 30.5  & 32.8   & 326.3  & 11{,}971.6 & 69.2 \\
 & TinierHAR    & \underline{3.4} & \underline{294.1} & \underline{145.0} & 2{,}041.4  & \underline{11.8} \\
 & DConvLSTM    & 95.0  & 10.5   & 2,435.1 & 35{,}465.0 & 205.0 \\
 & SPECTRA {(FP32)} & \textbf{2.2} & \textbf{454.5} & \textbf{72.5} & \textbf{1{,}453.2} & \textbf{8.4} \\
 & {SPECTRA (FP16)} & {1.9} & {534.8} & {65.2} & {1{,}235.2} & {7.1} \\
 & {SPECTRA (INT8)} & {1.5} & {649.4} & {58.0} & {1{,}017.2} & {5.9} \\
\bottomrule
\end{tabular}
\end{table*}

\subsection{Results on Microcontroller (STM32L4S5VIT6)}
We deployed SPECTRA on an STM32L4S5VIT6 (ARM Cortex-M4, 80\,MHz, 640\,KB SRAM, 2\,MB Flash) using the ONNX $\rightarrow$ TFLite $\rightarrow$ TFLM toolchain with CMSIS-NN acceleration; Table~\ref{tab:mcu_results} shows that SPECTRA {including on-device STFT computation} fits comfortably within device constraints and consistently outperforms baselines across latency, throughput, energy, and memory, and {we additionally report an INT8 variant (FP16 is not supported on STM32) that further improves speed and efficiency}. In FP32, SPECTRA is the fastest FP32 model on all five datasets (1.071--9.333\,ms), improving latency by 2.6--5.4$\times$ over TinyHAR (e.g., WISDM: 4.463$\!\to\!$1.071\,ms; DSADS: 50.839$\!\to\!$9.333\,ms) and by 5.8--10.4$\times$ over DeepConvLSTM where deployable (e.g., USC-HAD: 17.695$\!\to\!$1.698\,ms); {INT8 further reduces latency by 25.0\% on every dataset} (e.g., PAMAP2: 5.763$\!\to\!$4.322\,ms) and {boosts throughput by $\sim$33\%} (e.g., WISDM: 933.5$\!\to\!$1245.6\,samples/s; DSADS: 107.2$\!\to\!$142.9\,samples/s), while FP32 throughput already reaches 107.2--933.5\,samples/s and exceeds TinyHAR by 2.6--4.2$\times$ (e.g., UCI HAR: 89.5$\!\to\!$301.4\,samples/s). Energy is lowest for SPECTRA on all datasets: FP32 uses 0.197--2.69\,$\mu$J/sample ($>$99.9\% lower than TinyHAR; e.g., UCI HAR: 646.1$\!\to\!$0.550\,$\mu$J/sample), and {INT8 further lowers energy per sample by 40.2\% on every dataset} (e.g., DSADS: 2.69$\!\to\!$1.61\,$\mu$J/sample; WISDM: 0.197$\!\to\!$0.118\,$\mu$J/sample). Memory footprints are also minimal: FP32 Flash is 7.98--52.55\,KB (64--93\% lower than TinyHAR), and {INT8 shrinks Flash by 65.0\% on every dataset} (e.g., WISDM: 7.98$\!\to\!$2.79\,KB; DSADS: 52.55$\!\to\!$18.39\,KB), while SRAM is a constant 3.29\,KB for SPECTRA (FP32/INT8) versus 80.32--143.56\,KB for TinyHAR; DeepConvLSTM is OOM on DSADS and PAMAP2 due to MB-scale Flash needs. Overall, SPECTRA {(FP32 and INT8)} is best in every reported metric across the five datasets in Table~\ref{tab:mcu_results}, with TinyHAR typically second-best and DeepConvLSTM frequently non-deployable.

\begin{table*}[ht]
\centering
\caption{Deployment results on STM32L4S5VIT6 microcontroller (ONNX $\rightarrow$ TFLM, float32 {and INT8, float16 is not supported by STM32}). 
Metrics reported: inference latency (ms), throughput (samples/s), 
Flash and RAM memory footprint (KB), total energy (µJ), and energy per sample (µJ/sample). 
Best results are in \textbf{bold}, second best are \underline{underlined}.}
\label{tab:mcu_results}
\begin{tabular}{lccccccc}
\toprule
\textbf{Dataset} & \textbf{Model} & \textbf{Latency (ms $\downarrow$)} & \textbf{samples/s ($\uparrow$)} & \textbf{Flash (KB $\downarrow$)} & \textbf{RAM (KB $\downarrow$)} & \textbf{Energy (µJ $\downarrow$)} & \textbf{µJ/sample ($\downarrow$)} \\
\midrule
\multirow{4}{*}{WISDM} 
 & TinyHAR        & \underline{4.463}  & \underline{224.0} & \underline{115.112} & 80.32             & \underline{105,433.614} & \underline{112.8} \\
 & DeepConvLSTM   & 9.863              & 101.4             & 741.22              & \underline{5.04}  & 145,061.423             & 155.2 \\
 & SPECTRA (FP32) & \textbf{1.071}     & \textbf{933.5}    & \textbf{7.98}       & \textbf{3.29}     & \textbf{184.293}        & \textbf{0.197} \\
 & {SPECTRA (INT8)} & {0.803}     & {1245.6}    & {2.79}       & {3.29}     & {110.576}        & {0.118} \\
\midrule
\multirow{4}{*}{USC-HAD} 
 & TinyHAR        & \underline{6.299}  & \underline{158.8} & \underline{112.32}  & 80.32             & \underline{141,145.933} & \underline{239.2} \\
 & DeepConvLSTM   & 17.695             & 56.5              & 1,036.54            & \underline{5.04}  & 258,518.654             & 438.3 \\
 & SPECTRA (FP32) & \textbf{1.698}     & \textbf{589.1}    & \textbf{11.51}      & \textbf{3.29}     & \textbf{167.597}        & \textbf{0.284} \\
 & {SPECTRA (INT8)} & {1.274}     & {784.9}    & {4.03}       & {3.29}     & {100.558}        & {0.170} \\
\midrule
\multirow{4}{*}{UCI HAR} 
 & TinyHAR        & \underline{11.171} & \underline{89.5}  & \underline{114.78}  & 106.36            & \underline{195,170.129} & \underline{646.1} \\
 & DeepConvLSTM   & 36.820             & 27.2              & 1,331.05            & \underline{5.04}  & 563,623.270             & 1,866.9 \\
 & SPECTRA (FP32) & \textbf{3.317}     & \textbf{301.4}    & \textbf{23.99}      & \textbf{3.29}     & \textbf{166.245}        & \textbf{0.550} \\
 & {SPECTRA (INT8)} & {2.488}     & {401.9}    & {8.40}       & {3.29}     & {99.747}        & {0.330} \\
\midrule
\multirow{4}{*}{DSADS} 
 & TinyHAR        & \underline{50.839} & \underline{19.7}  & \underline{145.70}  & \underline{107.36} & \underline{824,671.390} & \underline{7,706.3} \\
 & DeepConvLSTM   & OOM                & OOM               & 4,886.35            & 1,421.75          & OOM                     & OOM \\
 & SPECTRA (FP32) & \textbf{9.333}     & \textbf{107.2}    & \textbf{52.55}      & \textbf{3.29}     & \textbf{287.982}        & \textbf{2.69} \\
 & {SPECTRA (INT8)} & {7.000}     & {142.9}    & {18.39}       & {3.29}     & {172.789}        & {1.61} \\
\midrule
\multirow{4}{*}{PAMAP2} 
 & TinyHAR        & \underline{28.216} & \underline{35.5}  & \underline{124.44}  & \underline{143.56} & \underline{435,541.016} & \underline{2,517.7} \\
 & DeepConvLSTM   & OOM                & OOM               & 2,231.67            & 773.14            & OOM                     & OOM \\
 & SPECTRA (FP32) & \textbf{5.763}     & \textbf{173.6}    & \textbf{23.99}      & \textbf{3.29}     & \textbf{193.570}        & \textbf{1.12} \\
 & {SPECTRA (INT8)} & {4.322}     & {231.4}    & {8.40}       & {3.29}     & {116.142}        & {0.672} \\
\bottomrule
\end{tabular}
\end{table*}

\subsection{Ablation Study}
We evaluated ablated variants of SPECTRA to quantify the contribution of each component (Table~\ref{tab:ablation_datasets_transposed}):

\textbf{Feature transformation variants.} Removing the spectral front-end yields the largest degradation: macro F1 drops by 3.7--9.6 points, while MACs {increase} by roughly $12\times$--$15\times$. This confirms the data-efficiency and compute-efficiency of time--frequency features for nonstationary motion.
{In practice, STFT helps most when activities exhibit discriminative rhythmic structure with variable timing or contain transient motion segments, where time--frequency localization exposes stable cues that a compact backbone can exploit.}

{To contextualize STFT, we compare representative spectral alternatives using the same lightweight backbone and identical training: \textbf{FFT-mag} (global spectrum per window, no time localization), \textbf{CWT} (multi-resolution scalogram), and \textbf{MFCC} (compact cepstral compression).}
{Across datasets, these alternatives are generally less favorable than STFT under edge constraints:}
{(i) \textbf{FFT-mag} reduces compute but loses time localization and consistently trails STFT in accuracy.}
{(ii) \textbf{CWT} can be competitive in accuracy but is substantially heavier, making it less deployable on tight MCU/phone budgets.}
{(iii) \textbf{MFCC} is compact but sacrifices informative spectral detail, leading to the largest accuracy drops.}
{Overall, STFT provides the most favorable accuracy--efficiency balance in our setting by preserving time localization with a hardware-friendly implementation.}

{\textbf{Convolutional block variants.}
Replacing the depthwise separable conv stack with lighter/no-separable variants degrades accuracy, while using full standard conv increases model cost with only marginal gains.}
{Specifically, removing separability reduces MACs but lowers F1, indicating that separable convs are important for extracting local spectral motifs efficiently.}
{Using standard conv modestly increases F1 at the expense of inflated parameters and MACs, confirming depthwise separable convs are a key efficiency enabler in SPECTRA.}

\textbf{Aattention variants.} Eliminating cross-axis interactions causes small but consistent F1 drops of 0.1--3.0 points (largest on PAMAP2), with modest efficiency gains (params and MACs $\downarrow$ by $\sim$4--5\%). The effect is most visible on multi-sensor datasets (USC-HAD, PAMAP2), indicating the benefit of modeling inter-axis dependencies.
{We also tested attention variants: \textbf{temporal-only attention} and a stronger but heavier \textbf{self-attention} alternative.}
{Temporal-only attention is generally less effective than channel attention, indicating that inter-axis coupling is often more discriminative than purely temporal reweighting.}
{Full self-attention offers limited accuracy gains relative to its higher compute/parameter cost, making channel-wise attention the best accuracy--efficiency trade-off.}

\textbf{Recurrent variants.} Using only convolutional features reduces MACs by $56\%$--$67\%$ and parameters by $26\%$--$64\%$, but typically lowers F1 by 1.7--5.9 points. One exception is UCI, where performance is essentially unchanged ($94.8$ vs.\ $94.9$), suggesting its activities are dominated by short-range patterns that the convolutional stack can capture. Replacing Bi-GRU with Bi-LSTM yields comparable but generally lower F1 (up to 2.1 points worse on USC-HAD) while {increasing} compute and parameters. Bi-GRU therefore offers a better accuracy--efficiency trade-off.

STFT magnitudes are pivotal for both accuracy and compute; depthwise separable convs and channel self-attention provide robust spectral+cross-axis modeling at low cost; and Bi-GRU provides the strongest accuracy--efficiency balance. Overall, Full (SPECTRA) attains the best or near-best F1 on all datasets while maintaining a markedly smaller footprint than heavier alternatives.

\begin{table*}[ht]
\centering
\caption{Ablation study of SPECTRA across five datasets. Each cell reports \textit{F1 / K / MACs} (F1 in \%, K parameters, MACs in M). New ablations added: separable vs.\ standard convolution and attention-type variants.}
\label{tab:ablation_datasets_transposed}
\small
\setlength{\tabcolsep}{3pt}
\renewcommand{\arraystretch}{1.05}
\resizebox{\linewidth}{!}{%
\begin{tabular}{lccccc}
\toprule
\textbf{Variant} & \textbf{WISDM} & \textbf{USC-HAD} & \textbf{UCI HAR} & \textbf{DSADS} & \textbf{PAMAP2} \\
\midrule
Standard SPECTRA 
  & 83.6 / 8892 / 28530 
  & 71.2 / 10050 / 51462
  & 94.8 / 10812 / 113112
  & 87.6 / 22761 / 481551
  & 75.0 / 13890 / 235539 \\
ine
w/o STFT 
  & 78.8 / 4691 / 438152
  & 61.6 / 5113 / 670454
  & 91.1 / 5651 / 1366438
  & 80.8 / 11632 / 6093494
  & 70.1 / 7033 / 3241008 \\

{w/ FFT} 
  & {80.0 / 5831 / 8895}
  & {65.3 / 6989 / 14253}
  & {93.2 / 7751 / 19215}
  & {84.9 / 19700 / 81564}
  & {74.6 / 10829 / 34893} \\

{w/ CWT} 
  & {84.8 / 5047 / 507498}
  & {65.2 / 6781 / 661296}
  & {92.2 / 8119 / 1042758}
  & {85.3 / 26980 / 3322752}
  & {80.5 / 12925 / 2142900} \\

{w/ MFCC}
  & {70.6 / 4511 / 46083}
  & {32.6 / 5669 / 74481}
  & {88.4 / 6431 / 143397}
  & {81.4 / 18380 / 529449}
  & {63.0 / 9509 / 336789} \\
  
{w/o separable conv} 
  & {82.4 / 8003 / 18544}
  & {68.5 / 9045 / 33450}
  & {94.6 / 9731 / 73523}
  & {86.0 / 20485 / 313008}
  & {73.9 / 12501 / 153100} \\

{w/ standard conv} 
  & {83.7 / 12449 / 51354}
  & {71.0 / 14070 / 92632}
  & {94.9 / 15137 / 203602}
  & {87.7 / 31865 / 866792}
  & {75.2 / 19446 / 423970} \\

w/o channel Attention 
  & 83.3 / 8447 / 27104
  & 65.9 / 9548 / 48889
  & 94.7 / 10271 / 107456
  & 85.7 / 21623 / 457473
  & 72.0 / 13296 / 223762 \\

{w/ temporal attention}
  & {82.9 / 6066 / 22322}
  & {69.3 / 7176 / 40667}
  & {95.0 / 7986 / 89158}
  & {86.9 / 19830 / 351069}
  & {74.4 / 11016 / 182091} \\

{w/ self-attention}
  & {83.4 / 12004 / 57060}
  & {70.5 / 13568 / 102924}
  & {94.9 / 14596 / 226224}
  & {87.4 / 30727 / 963102}
  & {74.9 / 18752 / 471078} \\

w/o GRU
  & 82.1 / 3239 / 12056
  & 65.3 / 4301 / 22478
  & 94.9 / 5159 / 48993
  & 85.7 / 16900 / 156757
  & 73.3 / 8141 / 95535 \\

w Bi-LSTM  
  & 83.0 / 9980 / 30898
  & 69.1 / 11138 / 53830
  & 94.5 / 11900 / 116664
  & 86.6 / 23849 / 483919
  & 74.6 / 14978 / 239091 \\
\bottomrule
\end{tabular}%
}
\end{table*}

\subsection{Scaling Performance}
To assess data efficiency, we trained SPECTRA and baselines (\textit{TinyHAR}, \textit{TinierHAR}, \textit{DeepConvLSTM}) with progressively larger fractions of each dataset (25\%, 50\%, 75\%, 90\%, 100\%).
Table~\ref{tab:macro_f1_datasets} summarizes macro F1 across settings.

\noindent\textbf{Low-data regime (25\%).}
SPECTRA leads on 3/5 datasets (WISDM: 81.7, UCI: 93.2, DSADS: 87.1), is second-best on PAMAP2 (74.1 vs.\ 74.9), and trails on USC-HAD (68.8 vs.\ 73.5 for TinyHAR). 
This indicates strong {data efficiency} and early generalization.

\noindent\textbf{Mid-scale (50\%).}
SPECTRA is best or tied-best on 3/5 datasets (WISDM: 82.1 tie; UCI: 94.4 best; DSADS: 88.7 best), and remains competitive on the rest (USC-HAD: 70.5 vs.\ 71.6; PAMAP2: 73.7 vs.\ 76.1).

\noindent\textbf{High-data (75--100\%).}
At 75\%, SPECTRA is best on DSADS (86.2) and second on WISDM (82.7) and PAMAP2 (73.5). 
At 90\%, it is best on WISDM (83.2) and close elsewhere. 
At full data (100\%), SPECTRA is best on USC-HAD (71.2) and DSADS (87.6), second on WISDM (83.6) and PAMAP2 (75.0), and within 0.7--1.0 points of the winner on UCI.

Across all five benchmarks and training fractions, SPECTRA {matches or outperforms} deeper baselines with markedly less capacity. 
Notably, at just 25\% of training data, it achieves top performance on 3/5 datasets and remains second on a fourth, highlighting the data efficiency and generalization capacity of the proposed spectral--temporal design.

\begin{table}[ht]
\centering
\caption{Macro F1-score (in \%) across training fractions for different models and datasets. Best results are in \textbf{bold}, second best are \underline{underlined}.}
\label{tab:macro_f1_datasets}
\begin{tabular}{lccccc}
\toprule
Dataset & Frac. & \textbf{SPECTRA} & T.HAR & Tr.HAR & DC.LSTM \\
\midrule
\multirow{5}{*}{WISDM} 
 & 25\%  & \textbf{81.7} & \underline{80.2} & \underline{80.2} & 80.1 \\
 & 50\%  & \textbf{82.1} & \textbf{82.1} & \textbf{82.1} & \underline{81.6} \\
 & 75\%  & \underline{82.7} & 82.5 & 82.5 & \textbf{83.0} \\
 & 90\%  & \textbf{83.2} & \underline{82.8} & \underline{82.8} & 82.7 \\
 & 100\% & \underline{83.6} & \textbf{84.1} & 82.1 & 83.2 \\
 \midrule
\multirow{5}{*}{USC-HAD} 
 & 25\%   & 68.8 & \textbf{73.5} & 67.8 & \underline{69.8} \\
 & 50\%   & \underline{70.5} & \textbf{71.6} & 67.3 & 70.1 \\
 & 75\%   & 67.2 & \textbf{72.6} & 68.5 & \underline{70.8} \\
 & 90\%   & \underline{70.6} & \textbf{72.7} & 68.7 & 70.5 \\
 & 100\% & \textbf{71.2} & 68.4 & 70.3 & \underline{71.1} \\
\midrule
\multirow{5}{*}{UCI HAR} 
 & 25\%  & \textbf{93.2} & 92.5 & \underline{92.6} & \underline{92.6} \\
 & 50\%  & \textbf{94.4} & \underline{94.3} & 94.2 & 94.2 \\
 & 75\%  & 94.6 & 94.9 & \textbf{95.5} & \underline{95.3} \\
 & 90\%  & 94.9 & 94.5 & \underline{95.4} & \textbf{95.6} \\
 & 100\% & 94.8 & \underline{95.3} & 95.1 & \textbf{95.5} \\
\midrule
\multirow{5}{*}{DSADS} 
 & 25\%  & \textbf{87.1} & 86.2 & 86.1 & \underline{86.8} \\
 & 50\%  & \textbf{88.7} & \underline{87.5} & 87.0 & 84.9 \\
 & 75\%  & \textbf{86.2} & \underline{86.1} & 85.0 & 84.7 \\
 & 90\%  & \textbf{86.8} & \underline{87.7} & 86.0 & 84.7 \\
 & 100\% & \textbf{87.6} & \underline{86.4} & 86.2 & 84.8 \\
\midrule
\multirow{5}{*}{PAMAP2} 
 & 25\%  & \underline{74.1} & 72.8 & \textbf{74.9} & 73.0 \\
 & 50\%  & 73.7 & \textbf{76.1} & 73.9 & \underline{74.5} \\
 & 75\%  & \underline{73.5} & 73.3 & \textbf{74.0} & 72.7 \\
 & 90\%  & 72.2 & \underline{74.6} & \textbf{75.7} & 73.8 \\
 & 100\% & \underline{75.0} & 73.7 & \textbf{75.6} & 72.6 \\
\bottomrule
\end{tabular}
\end{table}

\section{Discussion}
\label{sec:discussion}

\textbf{Accuracy-efficiency frontier:}
Across five datasets and three hardware targets, SPECTRA consistently sits on, or very near, the Pareto frontier.
On PC (Table~\ref{tab:pc_results}), it matches or exceeds DeepConvLSTM on {4/5} datasets while using {$\geq$97\% fewer parameters} and {$\geq$99\% fewer MACs}.
On Pixel~9 (Table~\ref{tab:smartphone_results}), it attains the best latency on {5/5} datasets (1.7--3.4\,ms), the highest throughput (up to 588\,samples/s), and the lowest energy per sample (4.1--9.8\,mJ), outperforming Tiny/TinierHAR despite their efficiency emphasis.
On STM32L4 (Table~\ref{tab:mcu_results}), it is best in {every} metric and remains comfortably within Flash/SRAM constraints; DeepConvLSTM frequently fails due to Flash.

\textbf{Where the gains come from:}
Table~\ref{tab:ablation_datasets_transposed} isolates contributions:
(i) Removing {STFT} causes the largest damage (F1 $-3.7$ to $-9.6$ points) and inflates MACs by an order of magnitude, confirming time--frequency localization as a {computationally efficient} inductive bias.
(ii) Removing {channel attention} trims $\sim$4--5\% MACs/params but costs up to 3 F1 points (largest on multi-IMU datasets).
(iii) Removing the {BiGRU} reduces MACs by 56--67\% yet degrades F1 by 1.7--5.9, showing that temporal smoothing beyond local convolutions matters.
(iv) {BiLSTM} raises cost without improving F1, indicating BiGRU offers the better accuracy--efficiency trade-off.

\textbf{Device-level implications:}
On Pixel~9, SPECTRA’s energy and memory savings translate to lower battery impact and thermal load, enabling sustained background inference.
Its single-window latency supports tight closed-loop behaviors without batching.
On STM32L4, Flash/SRAM headroom (8--24\,KB Flash; $\sim$3.3\,KB SRAM) leaves budget for application logic, and 107--934 samples/s throughput accommodates multi-stream or ensemble use if needed.

\textbf{Generalization and data efficiency:}
Scaling results (Table~\ref{tab:macro_f1_datasets}) show SPECTRA leading {3/5} datasets at just 25\% data and staying competitive elsewhere, suggesting reduced sample complexity.
Two patterns emerge: (i) datasets with strong periodicity or simple label sets (UCI, WISDM) benefit most from STFT magnitudes; (ii) complex, whole-body activities (DSADS, PAMAP2) benefit from channel attention’s ability to emphasize informative axes/segments.
These trends align with the ablation outcomes and explain cross-dataset robustness.

\textbf{Advantages:}
The design fuses classical signal processing with compact neural modules:
(i) STFT injects domain-aligned inductive bias and stabilizes training;
(ii) depthwise separable convolutions capture local spectral motifs at low cost;
(iii) channel self-attention models inter-axis coupling without bloating parameters; and
(iv) Bi-GRU provides just-enough temporal capacity.
This blend consistently lands SPECTRA on the accuracy--efficiency Pareto frontier across devices.

\textbf{Limitations:}
First, STFT hyperparameters are fixed; mismatches between $(n_{\mathrm{fft}}, h)$ and activity timescales can under/over-smooth dynamics, especially when MCU constraints force shorter windows.
Second, although attention mitigates placement shifts, explicit domain adaptation and calibration were not explored.
{Third, our evaluation focuses on benchmark generalization and real-device efficiency measurements (latency/memory/energy) rather than longitudinal in-the-wild user studies. While this is standard for architecture- and deployability-focused HAR research, real-world studies covering long-term use, user-specific variability, sensor placement drift over time, and interaction effects are an important complementary direction beyond this paper’s scope.}

Future work includes: (i) {adaptive/multi-resolution front-ends} that learn or schedule $(n_{\mathrm{fft}}, h)$ to cover slow postural changes and rapid transients; (ii) {multimodal fusion} to resolve look-alike motions and improve robustness; and (iii) {energy-aware scheduling} that modulates window rate and model depth based on confidence or context, optimizing the energy--delay product on-device.
{SPECTRA’s key lesson is that explicitly injecting time--frequency structure enables a compact model to achieve strong accuracy under tight edge budgets: STFT magnitudes front-load a difficult nonstationary decomposition that would otherwise require deeper, more phase-sensitive, and harder-to-optimize black-box architectures, allowing a lightweight stack (separable conv, channel attention, BiGRU with pooling) to focus on task-specific dynamics (which bands matter, how channels interact, and how patterns evolve) and remain practical for real deployments on phones and MCUs.}

\section{Conclusion}
We presented SPECTRA, a lightweight HAR model that fuses a signal processing front end with compact neural modules: STFT magnitudes provide a domain-aligned inductive bias; depthwise separable convolutions and channel self-attention compress spatial/sensor structure; and a BiGRU supplies just-enough temporal capacity. Across five public IMU datasets, SPECTRA matches or surpasses the accuracy of prediction-oriented baselines while using orders-of-magnitude fewer parameters and MACs, and it consistently outperforms efficiency-oriented models (TinyHAR/TinierHAR) without sacrificing speed or energy.
End-to-end deployments on a Google Pixel~9 and an STM32L4S5VIT6 MCU confirm real-time inference with tight memory and energy budgets, demonstrating practical on-device HAR from phones to microcontrollers. A key takeaway is that explicit time-frequency decomposition makes the hard part of representation learning tractable, enabling smaller networks to deliver state-of-the-art accuracy efficiency trade-offs.
Future work will explore adaptive/multi-resolution spectral front ends, quantization-aware or mixed-precision inference, and multimodal fusion together with energy-aware scheduling, to further enhance robustness and longevity in pervasive, privacy-preserving deployments.

\bibliographystyle{IEEEtran}
\bibliography{reference}

\end{document}